\documentclass[10pt,conference]{IEEEtran}
\usepackage{fancyhdr}
\usepackage{amsmath,amssymb}
\usepackage[pdftex]{graphicx}
\usepackage{multirow}
\usepackage{booktabs}
\usepackage{footnote}
\usepackage{threeparttable}
\ifCLASSINFOpdf
\else
\fi
\usepackage[caption=false,font=normalsize,labelfont=sf,textfont=sf]{subfig}

\begin{document}

%
\title{Detection Defense Against Adversarial Attacks with Saliency Map}

\author{\IEEEauthorblockN{Dengpan Ye, Chuanxi Chen\IEEEauthorrefmark{1}, Changrui Liu, Hao Wang, Shunzhi Jiang}
\IEEEauthorblockA{Key Laboratory of Aerospace Information Security and Trusted Computing,Ministry of Education,\\School of Cyber Science and Engineering,Wuhan University}
\{chencx\}@whu.edu.cn
}
\maketitle

\begin{abstract}
It is well established that neural networks are vulnerable to adversarial examples, which are almost imperceptible on human vision and can cause the deep models misbehave. Such phenomenon may lead to severely inestimable consequences in the safety and security critical applications. Existing defenses are trend to harden the robustness of models against adversarial attacks, e.g., adversarial training technology. However, these are usually intractable to implement due to the high cost of re-training and the cumbersome operations of altering the model architecture or parameters. In this paper, we discuss the saliency map method from the view of enhancing model interpretability, it is similar to introducing the mechanism of the attention to the model, so as to comprehend the progress of object identification by the deep networks. We then propose a novel method combined with additional noises and utilize the inconsistency strategy to detect adversarial examples. Our experimental results of some representative adversarial attacks on common datasets including ImageNet and popular models show that our method can detect all the attacks with high detection success rate effectively. We compare it with the existing state-of-the-art technique, and the experiments indicate that our method is more general.
\end{abstract}
\section{Introduction}
Deep Neural Networks (DNNs) have been demonstrated to perform exceptionally well in many fields, such as biometric identification (Krizhevsky et al. 2012), self-driving cars (Bojarski er al. 2016), malware detection (Dahl et al. 2013) and website traffic analysis (Zhang et al. 2019). However, adversarial attackers can force many DNNs based machine learning models to misclassify by adding small and imperceptible perturbations on original inputs to generate adversarial examples (Szegedy et al. 2014). Existing representative adversarial attack technologies usually utilize the key perturbation method: gradient based approach. Attackers could fast generate adversarial examples by using the gradient of loss function for deep network, then viewing implementing attack as an optimization problem and construct powerful adversarial examples with different distortion metric. Recent studies (Rozsa et al. 2016; Mirjalili et al. 2017; Wei et al. 2020;) show that adversarial examples can be implemented in physical scenarios, which gives rise to severe safety issues. Hence detecting or defending against adversarial examples has emerged as a critical factor in AI.

\par Almost as soon as the adversarial example attacks appeared, the researches of adversarial defense had been proposed. The deep network defense consists of two categories: active and passive defenses. The former means that a model can correctly classify adversarially perturbed images by hardening the network. Among many defense algorithms (Shafahi et al. 2019; Zhang and Wang 2019),adversarial training is the one state-of-the-art defense methods, in which the defender augments each minibatch of training data with adversarial examples. While the latter could be achieved by detecting and rejecting adversarial examples instead of modifying the models. This kind of defense is facilitating for depolying and applying though it is belong to the passive method. We also propose the detection defense against adversarial attacks along with the direction of this research. There are existing works aiming to implementing detection defenses. Xu et al. (2018) utilize the prediction inconsistency between the original and its modified image with special filters to implement detecting. Ma et al. (2019) propose to use the neural network invariant checking, i.e. analyzing the provenance channel and the activation value distribution channel to detect adversarial samples. Meng et al. (2017) suggests that it can train encoders and decoders to remove the added noises of the adversarial examples to further defend against attacks. Recently, a series of certified adversarial defenses have been proposed (Wong et al. 2018; Raghunathan et al. 2018; Zhang et al. 2018; Lecuyer et al. 2018). For example, Zhang et al. (2018) propose activation function as a certified defense, while Lecuyer et al. (2018) derive a better certified defense with differential privacy because a tight robustness guarantee is gained. However,these techniques are mostly applied to small networks and low-res datasets.
\par In this paper, we detect the potential adversarial examples in the model inputs to achieve the purpose of defense from the perspective of enhancing interpretability of deep networks. As we all known, the real subject contour could be observed by the thermal imager in the night. Inspired by this phenomenon, whether can we mark the main subject of image recognized by the deep networks with some technologies, so as to guide the model to explain its forecast for the object in the right direction. According to the recent researches for model interpretability (Simonyan et al. 2013; Springenberg et al. 2014; Zeiler and Fergus 2014; Zhou et al. 2016;  Selvaraju et al. 2020), if the activated regions of main features are different in the same image, the deep model would produce inconsistent predictive values for this image. On the other hand, it is generally necessary for attackers to construct the combination of specific adversarial noise to make the deep models misclassify in many scenarios (Athalye et al. 2018). Based on the two observations, we propose a novel defense method, the core of this defense method is considering the prediction inconsistency of the image by adopting the view of deep network interpretability. In other words, the model with the defense method will be able to determine whether the input contains potential adversarial perturbations by evaluating the prediction differences between the original input and the modified version containing activation characteristics.
\par We apply this novel defense to some classification tasks based on DNNs and evaluate it both on different attacks including gradient based attacks and frequently-used datasets including CIFAR-10 (Krizhevsky and Hinton 2009) and ImageNet (Russakovsky et al. 2015). The experimental results show that we can effectively detect the all attacks considered on the discussed datasets and models, with generally over $80\%$ detection accuracy. Besides, we compare our approach with state-of-the-arts detection approach, feature squeezing (Xu et al. 2018). The comparative results indicate that our proposed method can achieve high detection accuracy, and gain the more better prediction performance for the original inputs. The contributions of this paper are summarized as follows:
\par $\bullet$ We analyze the interpretability of model based on deep networks by using the saliency map for the activated region of image, which can highlight the key region of the image according to the gradient of the input;

\par $\bullet$ We propose a novel adversarial defense method to detect adversarial examples, whose key is utilizing that there is some inconsistency between the activation regions of the original input and that of the adversarial version, and there is integrity between the adversarial perturbations;
\par $\bullet$ We evaluate the effectiveness of our detection technique by testing the detection success rate against various attacks. The evaluation results show that this method can detect the all represent attacks considered with over $80\%$ accuracy. Comparing with the state-of-the-art detector, its detection success rate is almost on par with ours, while the proposed method can keep high prediction accuracy for the benign inputs.

\section{Background}
\subsection{Adversarial Examples and Attack}
Recent many researches show that neural networks are extremely vulnerable to adversarial examples. Essentially, an adversarial example is the input that is modified by adding the imperceptible adversarial noises. And the adversarial example is very similar to one of the original inputs correctly classified, but the deep models would give different prediction labels for the two inputs.
\par Consider a deep neural networks based classifier $f$ with parameters $\sigma$, an input $x$ with ground truth label $y$ and classification loss function $L$, while the $f$ could correctly classify the input, i.e., $f(x)=y$. The attackers usually seek for an adversarial example $x'$  such that the deep model classifies them differently. This is following the formulation:
\begin{equation}
\begin{split}
Find\ x',\\
s.t. f(x)\neq f(x'), \\
\Delta(x,x')< \epsilon.
\end{split}
\end{equation}
 Where $\Delta(x,x')$represents the difference between $x'$and $x$, $\epsilon$ is the adversarial manipulation budget. Generally, it is so difficult to solve eq (1) straightway to implement adversarial attacks, but the problem could be done by solving:
\begin{equation}
\max \limits_{\delta}L(x',y,\sigma),s.t. \Delta(x,x')< \epsilon.
\end{equation}
\par Actually, adversarial examples are initially demonstrated in (Szegedy et al. 2014; Biggio et al. 2013), have attracted great attention recently (Goodfellow et  al. 2015; Madry et al. 2018; Tramr et al. 2018; Biggio et al. 2017; Sun et al. 2020). Szegedy et al. (2014) pointed out that neural networks are vulnerable to adversarial examples and proposed an L-BFGS based algorithm to generate them. A fast gradient sign method (FGSM) for the adversarial attack generation is developed in Goodfellow et  al. (2015). Existing deep neural networks usually adopt some piecewise linear activation functions, which results in that any change of the original input is propagated to later hidden layers till the output layer for such deep models. Then, perturbations on inputs can accumulate in the propagated progress and lead to the misclassification. The representative FGSM attack is based on the above analysis and assumes the same attack strength at all dimensions. Formally, this kind of adversarial example is generated by using the following equation:
\begin{equation}
x'=x+\varepsilon \cdot sign(\nabla_{x}L(x,y,\sigma)),
\end{equation}
Where $\nabla$ represents the gradient and $L(\cdot )$ is the loss function used to train the model. Then, Kurakin et al. (2016) proposed an iterative version of FGSM, Basic Iterative Method (BIM). Recently, the projection gradient descent (PGD) attacks, a variant of BIM with uniform random noise as initialization is conducted
\begin{equation}
x^{t+1}=\mathcal{P}_{S}(x^t+\alpha \cdot sign(\nabla_{x}L(x^{t},y,\sigma))),
\end{equation}
where $\mathcal{P}_{S}(\cdot)$ is a projection operator projecting the input into the feasible region $S$, $\alpha T=\varepsilon$, $\alpha$ is the magnitude of the perturbation in each iteration and $T$ represents the iterations.

\par Besides, Carlini and Wagner proposed three different gradient based attacks with different $L_{p}$ norms(Carlini et al. 2017), namely, $L_{2}$, $L_{\infty}$ and $L_{0}$. In the most represent (Carlini\&Wagner) C\&W attack based on $L_{2}$ norm, there are three key innovations. Firstly, the adoption of logits $Z(\cdot)$ in the loss function, which hardens the adversarial attack against defensive distillation methods; Then, to gain the better trade-off between the prediction and distance values, they introduce an optimal constant ¦Á to control the adversarial example performance; Lastly, the key design in this attack is that it maps the target variable to the $argtanh$ space and the technique could reduce the dimension of inputs to decrease the computation cost of generating adversarial examples.
\par Powerful and representative adversarial attack methods are critical to better evaluating of defense techniques. The FGSM attack is one of the methods of fast generating adversarial examples, and is also the basic of many powerful PGD attacks. On the other hand, the C\&W attack based on $L_{2}$ norm has become the criterion of evaluating the feasibility of many defense methods. Consequently, if our proposed defense approach can effectively detect the two kinds of representative attacks, it is sufficient for evaluating the validity of this method.
\subsection{Adversarial Defense and Detection}
In the meantime, many efforts have been devoted to defend against adversarial examples (Xie et al. 2018; Guo et al. 2018; Samangouei et al. 2018; Song et al. 2017; Liao et al. 2018; Liu et al. 2020; Yang et al. 2020; Wang et al. 2020). Existing researches try to implement defense in three different categories: robust optimization, certified robustness and detection defense. In the robust optimization, various adversarial training methods (Goodfellow et al. 2015;  Madry et al. 2018; Tramr et al. 2018; Shafahi et al. 2019) are belong to the kind of  effective defense technologies against adversarial attacks. It can be traced back to Goodfellow et al. (2015), in which models are hardened by training on the sets with the adversarial examples. The robustness of models possessed by adversarial training relies on the attack strength of the different adversarial examples. In the certified robustness, a series of adversarial defenses have been proposed (Wong et al. 2018; Raghunathan et al. 2018; Zhang et al. 2018; Lecuyer et al. 2018). They generally seek a certified upper bound of the adversarial perturbations, the deep neural networks based model is supposed to be robust if the generating perturbations are less than this bound. Recently, Cohen et al. (2019) propose to adopt the randomized smoothing to construct a certified defense, while a better and tight certified robustness defense with additional noise is derived in Li et al. (2019).
\par Among the defense methods against adversarial examples, the detection based method is also one of the critical branches of the defense strategy, which is as important as producing correct output or establishing a certified upper bound. By use of the detection methods against adversarial attacks, the owner of model can find the potential adversarial examples and provide guidance for further denying or adjusting the inputs. There is one of the most popular detection method against adversarial examples: prediction inconsistency based algorithm. Tao et al. (2018) proposed to measure the inconsistency between original neural network and the enhanced network with some perceptible attributes to detect adversarial examples, the additional defined attributes are required for detection in this approach, which is the main limitation of this method due to the additional huge cost. Then, the state-of-the-art detection technique, feature squeezing achieves high detection rates for various attacks. The authors found that there is unnecessarily large input feature space in the deep neural networks, which allows an adversary to conveniently implement adversarial attacks. Thus, they reduced the redundant space by squeezing the input feature, i.e. reducing the bit depth of input image and spatial smoothness. They compared the difference between the prediction value of the original input and the squeezed sample, if the predicted value difference is greater than the set threshold, the input is considered to be adversarial. In this paper, we would compare with the feature squeezing defense to evaluate the validity of the proposed method based on the above discuss.
\par However, according to the work (Xu et al. 2018), the feature squeezing technique performance is not well on some attacks, such as FGSM and C\&W attacks on ImageNet set, while our experiments show that our proposed approach executes well against those attacks. Consequently, this may indicate that our work is the generalize of feature squeezing in a way.

\section{Design}
In this section, we first analyze the relation between the interpretability of deep networks based model and detection defense against adversarial examples. Then we discuss that how to implement adversarial detection under the perspective of interpretability. Lastly, we adopt some image processing technologies to harden our detection design.
\subsection{ Interpretability about Deep Networks}
Recently, the models based on deep networks have achieved unprecedented breakthroughs in different computer vision applications, such as image classification, semantic segmentation and object detection. However, these models are extremely vulnerable to adversarial examples, many scholars suggest that the lack of model interpretability may be the key reason that it is hard to defend against adversarial attacks. Hence, we make some discuss about the interpretability of deep networks for better implementing our detection.
\par Existing works (Simonyan et al. 2013; Springenberg et al. 2014; Zeiler and Fergus 2014; Zhou et al. 2016;  Selvaraju et al. 2020) usually regard the model interpretability as the model visualization problem. To explain the model outputs predicted by deep networks, they would highlight key pixels, whose changes in intensities could lead to the most impact on the final prediction score. Zeiler and Fergus (2014) proposed the classic Deconvolution method, it connects a deconvolution network behind the convolutional layer of the model and adopts the operation of transversal convolution kernel to map features to pixel space gradually, therefore the learned net of each layer in deep model is relatively transparent. Essentially, this method is based on gradient back propagation, which means that the gradient above zero is propagated forward. So, the main limitation of Deconvolution is that there are many noises in the visual image. To solve this problem, Guided backpropagation (Springenberg et al. 2014) combined the gradient with features by using ReLU, enabling the better reconstruction of the visual image with features assembled in the subject. However, despite producing fine-grained visual images, the two methods are not class discriminate, that means visual images about different classes are almost identic.
\par If we could not discriminate the categories in the model visualization, the interpretability of deep model is not sufficient for further deploying on other areas, and it is not convenient to apply to detecting adversarial examples. Zhou et al. (2016) found that Global Average Pooling (GAP) used by Network in Network  can not only prevent the deep networks from overfitting, but also generate feature map consistent with the number of original categories at the last convolutional layer containing the richest spatial and semantic information. Then, the Class Activation Mapping (CAM) can be obtained after weighting each channel in the feature map by using the gradient of the target category relative to the channel. A drawback of CAM is that it requires altering the model architecture and re-training, so it is only applicable to some particular kinds of deep networks which do not contain any fully-connected layers.
\par In order to apply the visualization of deep networks to our detection defense,
we consider utilize Gradient-weighted Class Activation Mapping
(Grad-CAM) method (Selvaraju et al. 2020). It also obtains the saliency map of the activated region by calculating the weighted sum of the feature map corresponding weights. But Grad-CAM uses the global average of the model gradient back propagation to calculate the weighted coefficients of the features instead of re-training the deep model. Broadly speaking, Grad-CAM combines feature maps using the gradient information that avoids making some modification in deep networks architecture. And it can be adopted in various of deep models. Consequently, it is feasible to apply the Grad-CAM to our detection defense without altering the target model architecture, and we can utilize this method to implement detection against different models.

\subsection{Adversarial detection}
Abstractly, deep network computation is essentially a process of taking a model input and producing the corresponding classification output. However, the models are usually vulnerable to the adversarial examples due to the uninterpretability of deep networks to some extent. Consequently, we implement detection defense against adversarial attacks from the view of the interpretability. Meanwhile, we destroy the integrity of adversarial perturbations to harden our defense by utilizing the additional noises.
\par Through the above discussion, the CAM can be used to explain that why the specific image is classified as the dog but not the cat in a deep model. Now, though the discussion of interpretability researches of deep network, we know that it is the saliency map that plays the key role in this progress, by gaining the saliency map of the discriminative activated region, the model could produce different results.
\par Therefore, our overarching idea is to utilize the saliency map to implement detection during deep model classification. The saliency maps are obtained from the Grad-CAM method, which combines with feature maps using the gradient information that avoids making some modification in deep networks architecture. Generally, the attack purpose is that he can change the prediction of a specific deep model, modifying the structure of the target model would be pointless and the cost would be unbearable. Similarly, when someone proposes a novel defense method against adversarial attacks, the case of keeping model architecture stable should be considered. Then, with saliency map contained activated features, we could take advantage of the prediction difference between the input and the  image produced by the saliency map to judge that whether the input sample contains potential threats. If the input is considered as a adversarial example, the prediction will be refused, otherwise the input is considered a benign sample.
\par Many works [Szegedy et al. 2014; Biggio et al. 2013; Goodfellow et al. 2015] suggest that adversarial perturbation could be accumulated due to the numerous net layers in the deep networks architecture.   And it is feasible for us to obtain the saliency map of the activated region of the input image sample without altering the deep networks model by the Grad-CAM. Hence, for the same input, there is the difference in the saliency maps of the activated region between the benign sample and adversarial example in the same trained deep networks the activation center of the benign must be the main critical part of the image on the human subjective visual, while adversarial example activation center could be in other region of the image.
\par \textbf{Activated region.} In order to find the desired activated area of input image in the deep network, we firstly generate the corresponding saliency map of this input by utilizing the Grad-CAM. In this strategy, the key weight $a_{k}^{c}$ with respect to the first $k$ feature map $f^{k}$ for the class $c$ is formulated by two steps:
\par\quad 1. Computing the gradient of the score for class $c$, $y_{c} $
$$
g_k=\frac{\partial y^{c}}{\partial f_{i,j}^{k}}
$$
\par\quad 2. Implementing the global average pooling for the  gradient
$$
a_{k}^{c}=\frac{1}{N}\sum_{i}\sum_{j}g_k
$$
Where the $N$ represents the number of pixels in the feature map, $y^{c}$ represents the logits output value about the class $c$ and $f_{i,j}^{k}$ refers to the activation at location $(i,j)$  of the feature map $f^{k}$.

\par Then, the critical saliency map $S_{c}$ could be obtained as follows:
\begin{equation}
 S_{c}=\sum_{k}a_{k}^{c}f^{k}
\end{equation}
Where the weight $a_{k}^{c}$ indicates the importance of every $f^{k}$ for the output class $c$. And the $S_{c}$ is essentially a weighted linear sum of the presence of the these feature map $f^{k}$. We gain the different weights $a_{k}^{c}$ by taking the gradient values, so it is sufficient to highlight the activated region most relevant to the particular class $c$.
\par Finally, in order to obtain the key region used for detection and avoid the disturbation of irrelevant pixels on the saliency map, ReLu function could be utilize to select pixels with positive excitation effect while calculating the saliency map. This can be expressed as:
\begin{equation}
 W^{c}=ReLU(S_{c})=ReLU(\sum_{k}a_{k}^{c}f^{k})
\end{equation}
\begin{figure}[h]
\centering
\includegraphics[width=8.5cm,height=8.5cm]{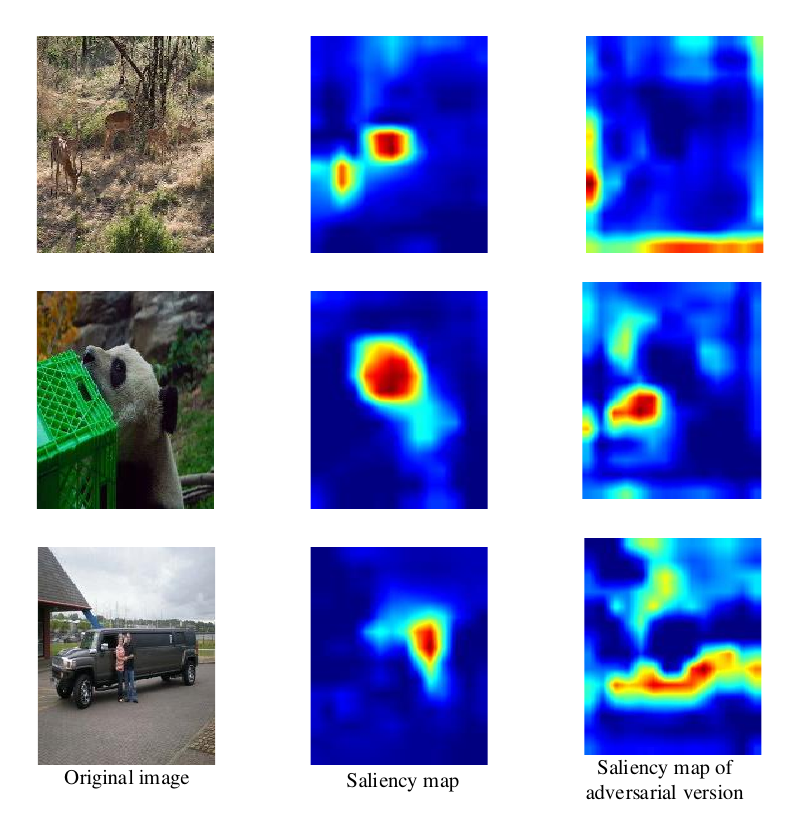}
\caption{The saliency maps of the original image and its adversarial example. they are established by the Grad-CAM visualization method, and there are some differences between the two kinds of saliency maps.}
\label{fig:label}
\end{figure}
\par \textbf{Detection implementation.} Now, we have obtained the saliency map of the input image activated region without altering the deep model with the (5) and (6).  Then, it is necessary to utilize the saliency map to implement defense against adversarial examples for the model. It is critical that how to find the distinctions of the saliency map of the activated region with respect to the input. An intuitive idea is to adopt the relation between the input and its saliency map, it is based on the following observations: Firstly, if the original input is correctly classified as a special category $c$, it is sufficient that the activated region of this image is concentrate upon the category $c$ by the saliency map; Secondly, adversarial perturbations are accumulated in the layers of deep networks, a small perturbation could lead to the difference of saliency map of adversarial example, The saliency maps of the original input and its adversarial version can be seen in Figure 1.
\par Then, in order to ensure that the activated region could play the role in the model input, we construct the novel image, named as subject emphasis image, which is generated by superimposing a certain proportion of the saliency map on the original input (see Figure 2). This design make it is convenient for predicting by ensuring consistency of the input format in deep networks.
\begin{figure}[h]
\centering
\includegraphics[width=8.5cm,height=3.2cm]{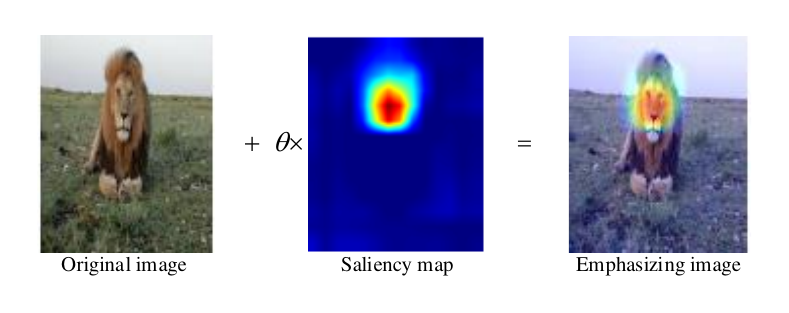}
\caption{The construction of the emphasizing image. The saliency map of the activated region is constructed by using visualization technology, then it overlaid on the original input with a proportion parameter $\theta$ to generate the emphasizing image.}
\label{fig:label}
\end{figure}
\par Therefore, after implementing the operation of obtaining the emphasis on the subject, the deep model should be more positive about the previous prediction class $c$ of the original input, that is, the prediction class of the original input is as the same as the those of the subject emphasis image. In other words, if the deep model gives an inconsistent class $t$ to the subject emphasis image, it is reasonable to judge that the input may be adversarial. In details, the generating of the saliency map utilizes the gradient of the input to generate the activated region according to (Selvaraju et al. 2020), which results in the activated region of the adversarial example is not stationary. Thus, the activated region is consistent to the main feature of the image that if it is benign. When the activated region of the image is variational, the corresponding classification label is different from that of the image, it is sufficient to ensure that this image is adversarial. The detection framework is shown as Figure 3.
\begin{figure*}[t]
\centering
\includegraphics[width=18cm,height=4.9cm]{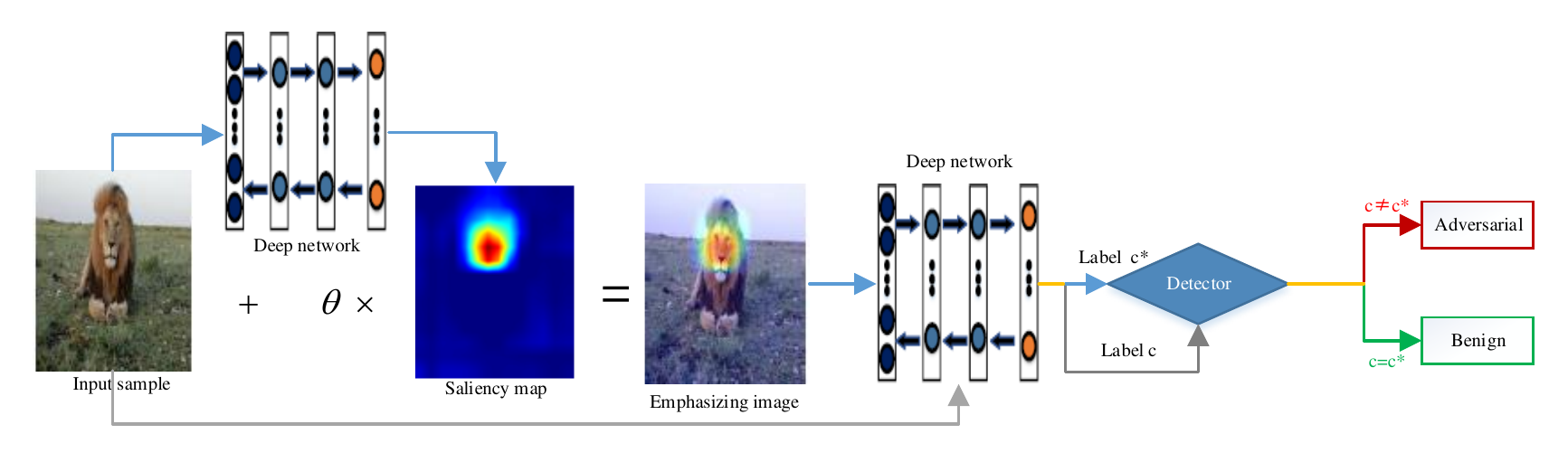}
\caption{The detection defense framework based on the inconsistency prediction. We can gain the key activated region (classification criteria) with the emphasizing image, and the activated region of the input with adversarial perturbation is generally different from the that of the benign input. Then, by comparing the output labels that are corresponding to the input and the relative emphasizing images to detect the adversarial examples.}
\label{fig:label}
\end{figure*}
\par It is notable that there is a hyper-parameter $\theta$ in the detection framework. To some extent, the parameter $\theta$ play the key role in the detection progress. Following the previous process mode in the deep learning community, we set the value range of $\theta$ is 0.1$\sim$0.3 according to the experimental operations.
\subsection{Enhanced detection}
Generally, there is the integrity in adversarial example of the original input, it could result in the false consistency of the saliency map that makes the high false nagative rate of the detection sometimes. In this section, we implement disturbation to the adversarial examples with the additional noises, which can disturb the relevance between some pixels and different regions in the adversarial image. Consequently, it can highlight the difference of the saliency map to enhance the detection.
\par \textbf{color reversing.} Different from grayscale images, there are RGB three channels in natural images, different colors can be represented by the combination of different values of each channel. In practice, it is the contour and the critical feature (ear, fur or limbs) play the key role in identifying objects (dog or cat). Even wearing red and blue glasses, we also can accurately identify the dog or cat. This indicates that the object color may work for auxiliary judgment in our identification. Meanwhile, we find that the activated region of the model on the image is also in the key position of the object to be predicted in the deep networks model. Hence, under the premise that the main contour of the object remains unchanged, the change of the image overall color only leads to a minimal impact on the activated region. However, it is the accumulation of small adversarial perturbations that eventually cheats the model in adversarial attacks. On the other hand, there are three dimensions in RGB mode for the convolution kernel of model, and the corresponding value of each convolution kernel is not the same. So, we can make the conversion from RGB to BGR in source to disturb the accumulation of perturbations, which breaks the integrity of adversarial examples in some degree. That is, it can strengthen the inconsistency of saliency map without influencing the activated region to make the detection defense credible.
\par \textbf{Zero mean.} The brightness value of image is also the auxiliary factor for identifying an object, in other words, the change of brightness value of the image will not produce much impact for identification. Based on this observation, we can alter brightness value with zero mean method, which keeps the main contour of the original input, but disturbs the integrity of adversarial examples. Further, the strategy of zero mean can reduce the similar parts between the test images and the whole training set to highlight the key differences, it is also favorable for detecting  adversarial examples. In details, we subtract the values of the natural image three channels by the average of all pixel points of the corresponding channels on the training set. Then, we remove the average brightness value of the input test samples for implementing efficiently in this paper. The experimental results show that the zero mean can improve the detection accuracy.These processed images are shown in  Figure 4.
\begin{figure}[h]
\centering
\includegraphics[width=8.5cm,height=3.2cm]{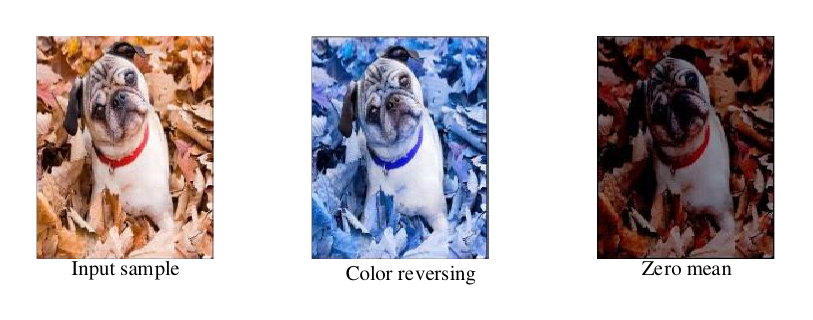}
\caption{The image process with additional noise. The operations of color reversing and zero mean are almost not interferential to identify object for deep network that is pay more attention to the contour and the key features of object.}
\label{fig:label}
\end{figure}

\section{Evaluation}
In this section, we discuss the results of some scale evaluation. On the one hand, we implement the proposed detection defense against the mainstream attack algorithms on different deep network models to verify the feasibility of the method. On the other hand, we compare this method with the state-of-the-art feature squeezing technology, the experimental results show that the  performance of our method is at least on par with it, and is better than it at other cases, while the proposed method can ensure the high prediction accuracy for the original inputs.

\subsection{Setup}
\par \textbf{Datasets.} For mainstream attack algorithms, we perform our experiments on two popular image datasets: CIFAR-10  and ImageNet . They are colored image datasets used for image classification. For the ImageNet dataset, we use the ILSVRC2012 samples, which contains more than 1.2 million training set image samples and their labels, 50,000 verification set image samples and 100,000 test set images. However, for the CIFAR-10 dataset, the image size of this set is only $32*32$, which results in poor visualization and is difficult to detect in this defense. Besides, the number of categories in the set does not meet our expectations, so the results for this dataset are not presented in this paper. We chose these datasets because they are the most widely used datasets for classification task and most representative attacks are carried out on them.
\par \textbf{Models.} We selected the VGG16 model, ResNet50 model as the main experimental model, these models are representatives of their kinds and used in the attacks under study (Goodfellow et al. 2015; Tramr et al. 2018; Carlini et al. 2017). At the same time, in order to make a comparison between different defense, we also evaluated on the relevant model. In model construction, it is necessary for training ImageNet to require expensive hardware equipment and vast time, so we choose the pre-training model Keras on ImageNet. But we still fully reserve the initialization process of parameters in the original model, so as to maintain the prediction accuracy of the model.
\par \textbf{Attack.} We evaluated our detection method on the most representative attacks described in background section. For the untargeted gradient based attacks, we considered the FGSM attack and used the implementations from the ART library (Nicolae et al. 2019) to generate adversarial examples. On the other hand, for the targeted gradient based attacks (three C\&W attacks), we adopted implementations from the Carlini et al. (2017) and focused on the C\&W attack based on $L_{2}$ norm, i.e., C\&W${_2}$ attack. Besides, it is worth noting that we selected the attack parameters could be optimized in implementing attacks, so as to make the generated adversarial examples possess high credibility as far as possible. In the meantime, we reused the same attack parameters for the Feature Squeezing to make the better comparison objectively.

\subsection{Detecting attacks with saliency map}
This section shows the experimental results about detecting adversarial examples with saliency map. We adopt the subject emphasis image, which is generated by superimposing a certain proportion of the saliency map on the original input. For the detection direction of adversarial examples, the detection method should have the following two abilities. When the input image is the benign, the proposed detection method cannot significantly affect the predicted value by deep networks model, especially cannot change the original prediction category of the image (ImageNet is top-1 accuracy); While the input image is the adversarial, the detection method can effectively detect the illegal input, and make early warning and prevent the input to the deep network model according to the defense strategy. We conducted experiments on different deep network models, and the experimental results are shown in Table 1.
\begin{table}[!h]

\scriptsize

\centering

\caption{The detection effects of the models with saliency map}

\label{Tab03}
\begin{threeparttable}
\begin{tabular}{ccccccc}

\toprule

\multirow{2}{*}{Proportion ($\theta$)} & \multicolumn{3}{c}{VGG16} & \multicolumn{3}{c}{ResNet50} \\

\cmidrule(r){2-4} \cmidrule(r){5-7}

&  OSPA$^{1}$   &  FGSM   &   C\&W${_2}$

&  OSPA    &  FGSM   &   C\&W${_2}$

 \\

\midrule

0.0      & 100.0          & 0.0         & 0.0
                  & 100.0           & 0.0         & 0.0       \\

0.1    & 94.8           & 18.4        & 30.4
                  & 85.2           & 22.2       & 27.2    \\

0.2     & 84.5          & 42.0       & 55.1
                  & 70.4           & 47.3       & 48.9    \\

0.3    & 78.2          & 53.5       & 63.8
                  & 51.7          & 66.5       & 64.1     \\

0.4    & 73.0          & 61.0      & 72.5
                  & 37.9          & 74.4       & 73.9    \\

0.5    & 69.5         & 68.4       & 76.8
                  & 23.7          & 83.3       & 77.2     \\

\bottomrule

\end{tabular}
\begin{tablenotes}
        \footnotesize
        \item[1] It represents the Original Samples Prediction Accuracy (OSPA) on the model, its meaning is the same in the following.
      \end{tablenotes}
    \end{threeparttable}
\end{table}
\par As shown in Table 1, when the saliency map proportion is 0 in the subject emphasis image, it is equivalent to there is no defense measures, so the detection rate for adversarial examples is $0\%$, while the prediction accuracy for the corresponding original inputs is $100\%$. Further, with the increase of saliency map proportion, the detection success rate for adversarial examples increases gradually, and the prediction accuracy for benign samples decreases.
For the VGG16 model, when the saliency map proportion paramater $\theta$ is equal to $0.5$, the detection success rate against FGSM, C\&W$_{2}$ attacks can achieve $68.4\%$, $76.8\%$ respectively. For the ResNet 50 model, the corresponding detection success rate against FGSM, C\&W$_{2}$ attacks can reach maximum $83.3\%$, $77.2\%$. However, the prediction accuracy for the original inputs decreases in those situations, it suggests that the detection defense based on inconsistency with saliency map sometimes is not enough. We combine with other proposed methods to improve this situation and more details will be described in the subsequent sections.

\subsection{Detecting attacks with saliency map \& color reversing}
As we have analyzed in design section, color reversing of image does not have a particularly significant impact on human identification, because we mostly use the contour of the subject for distinguishing. deep networks also adopt this feature to some extent, that is, it is not very sensitive to color reversing. However, the adversarial examples are misclassified through the accumulation of errors, so the color reversing of the three channels in RGB images will lead to generate the activation offset of the adversarial examples at each layer. It breaks the integrity of adversarial examples, which can strengthen the inconsistency of saliency map between the original input and the adversarial example. So, we evaluated saliency map combined with this method, and the results are shown in Table 2.
\begin{table}[!h]

\scriptsize

\centering

\caption{The detection effects of the models with saliency map \& color reversing}

\label{Tab03}
\begin{tabular}{ccccccc}

\toprule

\multirow{2}{*}{Proportion ($\theta$)} & \multicolumn{3}{c}{VGG16} & \multicolumn{3}{c}{ResNet50} \\

\cmidrule(r){2-4} \cmidrule(r){5-7}

&  OSPA   &  FGSM   &   C\&W${_2}$

&  OSPA    &  FGSM   &   C\&W${_2}$

 \\

\midrule

0.0      & 94.8          & 70.7         & 53.6
                  & 92.1           & 73.4         & 54.4       \\

0.1     & 94.3           & 69.5        & 47.8
                  & 88.7           & 75.9       & 47.8    \\

0.2     & 91.4          & 69.5       & 63.8
                  & 80.8           & 80.3       & 64.1    \\

0.3   & 89.1         & 77.0       & 68.1
                  & 64.0          & 86.7       & 69.6     \\

0.4    & 81.0          & 78.2      & 75.4
                  & 46.3          & 91.1      & 72.8    \\

0.5    & 78.7         & 81.0       & 79.7
                  & 27.1          & 93.1       & 77.2     \\

\bottomrule

\end{tabular}

\end{table}
\par It can be found from Table 2 that the saliency map with color reversing, as we have pointed, effectively detects the adversarial examples for different attacks, while ensures the prediction accuracy for the original inputs in deep networks models. More specifically, for the VGG16 model, when only the color reversing method is used to detect the adversarial examples, the detection success rate is more than $70\%$, and the prediction accuracy for the original inputs declines by $5.2\%$. However, this image preprocessing defense of the color reversing is not robust. If the attacker access to the strategy of color reversing, the adversarial examples could be generated in a different color space to evade detection. Thus, it is sufficient to combine the saliency map with color reversing for robust detection. Then, as shown in Table 2, when the saliency map proportion is $50\%$, the detection success rate against FGSM, C\&W$_{2}$ attacks can achieve $81.0\%$, $79.7\%$ respectively for the VGG16 model. For the ResNet 50 model, the corresponding detection success rate against FGSM, C\&W$_{2}$ attacks can reach maximum $93.1\%$, $77.2\%$.

\subsection{Detecting attacks with saliency map \& zero mean}
In this section, we evaluated the detection effect of saliency map combined with zero mean. As we all known, the brightness value of image is the subordinate factor for identifying an object. So, the zero mean method for altering brightness value could not interfere the effort of the saliency map, while the integrity of adversarial examples could be disturbed. And the experiments are carried out on the prediction accuracy for the original images and the detection success rate for adversarial examples. The experimental results on the different deep models are shown in Table 3.

\begin{table}[!h]

\scriptsize

\centering

\caption{The detection effects of the models with saliency map \& zero mean}

\label{Tab03}
\begin{tabular}{ccccccc}

\toprule

\multirow{2}{*}{Proportion ($\theta$)} & \multicolumn{3}{c}{VGG16} & \multicolumn{3}{c}{ResNet50} \\

\cmidrule(r){2-4} \cmidrule(r){5-7}

&  OSPA   &  FGSM   &   C\&W$_{2}$

&  OSPA    &  FGSM   &   C\&W$_{2}$

 \\

\midrule

0.0      & 93.7          & 82.8         & 63.8
                  & 92.6           & 88.2         & 68.5       \\

0.1     & 93.7           & 82.8        & 69.6
                  & 93.6           & 89.2       & 72.8    \\

0.2     & 92.5          & 82.2       & 73.9
                  & 92.1           & 89.7       & 73.9    \\

0.3    & 86.8         & 82.2       & 76.8
                  & 85.7          & 90.2       & 77.2     \\

0.4   & 81.0          & 83.9      & 76.8
                  & 80.8          & 89.2      & 78.3    \\

0.5     & 78.2        & 85.1       & 78.3
                  &73.9          & 90.2      & 78.3     \\

\bottomrule

\end{tabular}

\end{table}
\par As shown in Table 3, the saliency map combined with zero mean method can also effectively detect the adversarial examples for different attacks, while ensures the prediction accuracy for the original inputs in DNNs models. In other words, For the VGG16 model, when the saliency map proportion paramater $\theta$ is equal to $0.3$, the detection success rate against FGSM, C\&W$_{2}$ attacks can achieve $82.2\%$, $76.8\%$ respectively. And for the ResNet 50 model, the corresponding detection success rate against FGSM, C\&W$_{2}$attacks can reach maximum $90.2\%$, $78.3\%$ respectively, while the prediction accuracy for the original is about more than $73\%$.

\subsection{Comparison}
In order to show the effectiveness of the proposed defense method intuitively,
we compared it with the state-of-the-art method, named as feature squeezing. The author found that some unnecessary feature input space brings great convenience for the attacker to construct adversarial examples, so it reduced the redundant space by squeezing the input feature to implement defending. Besides, we first reproduce the feature squeezing on the adversarial attack and defense platform ART. Then, for the median filtering of this method,
the ndimage module in Scipy library was used, and the open CV library was adopted for the non-local mean filtering. The experimental results of defending against FGSM attack on VGG16 model by Feature Squeezing are shown in Table 4

\begin{table}[t]

\scriptsize

\centering

\caption{The detection effects for VGG16 by feature squeezing}

\label{Tab03}
\begin{tabular}{c|ccc}

\hline{methods} &{parameters} &{OSPA} &{FGSM}\\
\hline
\multirow{6}{*}{Image bit depth}
& {1-bit} & {13.0} & {8.6}    \\

& {2-bit} & {55.8} & {24.1} \\

& {3-bit} & {84.2} & {20.7}  \\

& {4-bit} & {94.0} & {2.9}  \\

& {5-bit} & {97.7} & {3.5}  \\
& {6-bit} &{99.5} &{0.6}\\
\hline

\multirow{2}{*}{Median filtering}
& {2*2} & {86.1} & {20.1}  \\

& {3*3} & {76.7} & {21.8} \\
\hline

\multirow{4}{*}{Nonlocal mean filtering}
& {11-3-2} & {96.3} & {31.0} \\
& {11-3-4} & {90.7} & {36.8} \\
& {13-3-2} & {96.7} & {32.2} \\
& {13-3-4} & {91.2} & {36.7} \\
\hline
Optimal combination &{5-bit,2*2,11-3-4} & {78.6} & {50.6}\\
\hline

\end{tabular}

\end{table}
\par As shown in Table 4, in the image bit depth, the parameter 8-bit represents that no modification has been made to the input, so the prediction accuracy for the inputs is $100\%$ and the detection success rate for the adversarial examples is $0\%$. When we make the tradeoff between the prediction accuracy for the original inputs and the detection success rate for the adversarial examples, the better different squeezing parameters are 5-bit, $2\* 2$ and 11-3-4 respectively in the table. And we can find that the utilization of the combination of various squeezing method in Feature Squeezing can achieve better defense effect (see last line in table 4). Next, we compared the optimal combination of our proposed defense strategy with the feature squeezing defense method. Although there is a better detection success rate in the proposed strategy with saliency map, we still combine it with color reversing or zero mean. Not only is the thin point of defense not a good choice, but also could the latter two technologies break the integrity of adversarial examples to highlight the inconsistency between the original and adversarial inputs. In other words, on the basis of utilizing saliency map, which is inspired by the view of enhancing the interpretability of deep networks, the defense point of color reversing or zero mean is added in the proposed defense strategy to ensure its optimal performance. To make the evaluation more objective, the experiments are based on the VGG16 model and the results are shown in Table 5.
\begin{table}[!h]

\scriptsize

\centering

\caption{The comparison of detection effects between our method and feature squeezing}

\label{Tab03}
\setlength{\tabcolsep}{1mm}{
\begin{tabular}{ccccc}

\toprule

{} &{Optimal defense parameters}&OSPA &FGSM &C\&W$_{2}$\\

\midrule

{Feature Squeezing}  & {5-bit,2*2,11-3-4}  & 78.6  & 50.6   & 91.5    \\

{saliency map \& color reversing}  & {$\theta=0.1$} & \textbf{94.3} & \textbf{69.5}   & \textbf{47.8}    \\

{saliency map \& zero mean}    & {$\theta=0.1$}  & \textbf{93.7}   & \textbf{82.8}    & \textbf{69.6}    \\

\bottomrule

\end{tabular}}

\end{table}
\par It can be found from the Table 5, when the saliency map proportion added  is equal to $0.1$, and the additional color reversing or zero mean is combined, for the FGSM attack, the detection success rate of our proposed defense method is $69.5\%$ or $82.8\%$, which is more than the detection success rate of feature squeezing method. For the C\&W$_{2}$ attack, the prediction accuracy for the original inputs of our method is $93.7\%$ or $94.3\%$, it is still higher than those of feature squeezing method, and the detection effect of the proposed method is also well. According to the work (Xu et al. 2018), this is possible that there is a limitation of feature squeezing: requiring high quality squeezers for various attacks. Hence, our approach is more general compared with it.
\section{Conclusion}
In this paper, we present a novel prediction inconsistency based detection algorithm against adversarial examples, which is based on the view of enhancing model interpretability: taking advantage of the saliency map of the activated region to visual the prediction process by the deep networks. We then construct the defense framework with the saliency map combined with additional noise techniques to implement detection. Our evaluation results show that our method can accurately detect current representative attacks, and the performance of proposed method is at least on par with the state-of-the-art technique,and is better than it at other cases. However, there are still a lot to explore along this research theme. Such as, is there another better interpretability view of deep networks? How to apply the algorithm to defend against attack in black-box scenarios? We hope that these questions are addressed very well in the future work.
\section*{Acknowledgment}
 The views and conclusions contained herein are those of the authors and should not be interpreted as necessarily representing the official policies or endorsements.

\section{References}

\smallskip \noindent
Krizhevsky, A., Sutskever, I., and Hinton, G. 2012. Imagenet classification with deep convolutional neural networks. \textit{In advances in neural information processing systems}, 1097--1105.

\smallskip \noindent
Bojarski, M.; Deltesta, D.; Dworakowski, D.; Firner, B.; Flepp, B.; Goyal, P.; Jackel, L. D.; Monfort, M.; Muller, U.; and Zhang, J. 2016. End to end learning for self-driving cars. \textit{CoRR} abs/1604.07316.

\smallskip \noindent
Dahl, G. E.; Stokes, J. W.; Deng, L.; and Yu, D. 2013. Large-scale malware classification using random projections and neural networks. \textit{In Acoustics, Speech and Signal Processing (ICASSP)}.

\smallskip \noindent
Zhang, X. K.; Hamm, J. H.; Reiter, M. K.; and Zhang, Y. Q. 2019. Statistical privacy for streaming traffic. \textit{In proceedings of the Network and Distributed Systems Security Symposium (NDSS)}.

\smallskip \noindent
Szegedy, C.; Zaremba, W.; Sutskever, I.; Bruna, J.; Erhan, D.; Goodfellow, I.; and  Fergus, R. 2014. Intriguing properties of neural networks. \textit{CoRR} abs/1312.6199.

\smallskip \noindent
Rozsa, A.; Gunther, M.; Rudd, E. M.; and Boult, T. E. 2016. Are facial attributes adversarially robust? \textit{In International Conference on Pattern Recognition (ICPR)}.

\smallskip \noindent
Mirjalili, V.; and Ross, A. 2017. Soft biometric privacy: Retaining biometric utility of face images while perturbing gender. \textit{In International Joint Conference on Biometrics (IJCB)}.

\smallskip \noindent
Shafahi, A.; Najibi, M.; Ghiasi, A.; Xu, Z.; Dickerson,J.; Studer, C.; Davis£¬L. S.; Taylor, G.; and Goldstein, T. 2019. Adversarial training for free! \textit{In the Conference on Neural Information Processing Systems (NeurIPS)}.

\smallskip \noindent
Zhang, H. C.; and Wang, J. Y. 2019. Defense against adversarial attacks using feature scattering-based adversarial training. \textit{In the Conference on Neural Information Processing Systems (NeurIPS)}.

\smallskip \noindent
Xu, W.; Evans, D.; and Qi, Y. 2018. Feature Squeezing: Detecting adversarial examples in deep neural networks. \textit{In proceedings of the Network and Distributed Systems Security Symposium (NDSS)}.

\smallskip \noindent
Ma, S. Q.; Liu, Y. Q.; Tao, G. H.; Lee, W. C.; and Zhang, X. Y. 2019. NIC: Detecting adversarial samples with neural network invariant checking. \textit{In proceedings of the Network and Distributed Systems Security Symposium (NDSS)}.

\smallskip \noindent
Meng, D.; and Chen, H. 2017. Magnet: a two-pronged defense against adversarial examples. \textit{In proceedings of the ACM SIGSAC Conference on Computer and Communications Security (CCS)}.

\smallskip \noindent
Wong, E.; Schmidt, F.; Metzen, J. H.; and Kolter, J. Z. 2018. Scaling provable adversarial defenses. \textit{In the Conference on Neural Information Processing Systems (NeurIPS)}.

\smallskip \noindent
Raghunathan, A.; Steinhardt, J.; and Liang, P. S. 2018. Semidefinite relaxations for certifying robustness to adversarial examples. \textit{In the Conference on Neural Information Processing Systems (NeurIPS)}.

\smallskip \noindent
Zhang, H.; Weng, T. W.; Chen, P. Y.; Hsieh, C. J.; and Daniel, L. 2018. Efficient neural network robustness certification with general activation functions. \textit{In the Conference on Neural Information Processing Systems (NeurIPS)}.

\smallskip \noindent
Lecuyer, M.; Atlidakis, V.; Geambasu, R.; Hsu, D.; and Jana, S. 2018. Certified robustness to adversarial examples with differential privacy. \textit{CoRR} abs/1802.03471.

\smallskip \noindent
Athalye, A.; Carlini, N.; and Wagner, D. 2018. Obfuscated gradients give a false sense of security: Circumventing defenses to adversarial examples. \textit{CoRR} abs/1802.00420.

\smallskip \noindent
Krizhevsky, A.; and Hinton, G. 2009. Learning multiple layers of features from tiny images. Technical report.

\smallskip \noindent
Russakovsky, O., Deng, J., Su, H., Krause, J., Satheesh, S., Ma, S., Huang, Z., Karpathy, A., Khosla, A., Bernstein, M., Berg, A. C., and Li, F. F. 2015. Imagenet large scale visual recognition challenge. \textit{International Journal of Computer Vision} 115(3): 211--252.

\smallskip \noindent
Biggio, B.; Corona, I.; Maiorca, D.; Nelson, B.; Srndic, N.; Laskov, P.; Giacinto, G.; and Roli, F. 2013. Evasion attacks against machine learning at test time. \textit{In European Conference on Machine Learning and Principles and Practice of Knowledge Discovery in Databases (ECML PKDD)}

\smallskip \noindent
Wei, Z. P.; Chen, J. Q.; Wei, X. X.; Jiang, L. X.; Jiang, Y. G.; Chua, T. S.; ang Zhou, F. F. 2020. Heuristic Black-box Adversarial Attacks on Video Recognition Models. \textit{In Proceedings of the AAAI Conference on Artificial Intelligence (AAAI)}.

\smallskip \noindent
Goodfellow, I.; Shlens, J.; and Szegedy, C. 2015. Explaining and harnessing adversarial examples. \textit{In International Conference on Learning Representations (ICLR)}.

\smallskip \noindent
Madry, A.; Makelov, A.; Schmidt, L.; Tsipras, D.; and Vladu, A. 2018. Towards deep learning models resistant to adversarial attacks. \textit{In International Conference on Learning Representations (ICLR)}.

\smallskip \noindent
Sun, J. W.; Zhang, T. W.; Xie, X. F.; Ma, L.; Zheng, Y.; Chen, K. J.; and Liu, Y. 2020. Stealthy and Efficient Adversarial Attacks against Deep Reinforcement Learning. \textit{In Proceedings of the AAAI Conference on Artificial Intelligence (AAAI)}.

\smallskip \noindent
Tram¨¨r, F.; Kurakin, A.; Papernot, N.; Boneh, D.; and McDaniel, P. 2018. Ensemble adversarial training: Attacks and defenses. \textit{In International Conference on Learning Representations (ICLR)}.

\smallskip \noindent
Biggio B.; and Roli, F. 2017. Wild patterns: Ten years after the rise of adversarial machine learning. \textit{CoRR} abs/1712.03141, 2017.

\smallskip \noindent
Kurakin, A.; Goodfellow, I.; and Bengio, S. 2016. Adversarial examples in the physical world. \textit{CoRR} abs/1607.02533, 2016.

\smallskip \noindent
Carlini, N.; and Wagner, D. 2017. Towards evaluating the robustness of neural networks. \textit{In IEEE Symposium on Security and Privacy (SP)}.

\smallskip \noindent
Xie, C.; Wang, J.; Zhang, Z.; Ren, Z.; and Yuille, A. 2018. Mitigating adversarial effects through randomization. \textit{In International Conference on Learning Representations (ICLR)}.

\smallskip \noindent
Guo, C.; Rana, M.; Ciss, M.; and Vandermaaten, L. 2018. Countering adversarial images using input transformations. \textit{In International Conference on Learning Representations (ICLR)}.

\smallskip \noindent
Shafahi, A; Najibi, M.; Xu, Z.; Dickerson, J.; Davis, L.; and Goldstein, T. 2019. Universal Adversarial Training. \textit{In Proceedings of the AAAI Conference on Artificial Intelligence (AAAI)}.

\smallskip \noindent
Wang, C. Y.; He, X. F.; and Zhou, A. Y. 2019. Improving Hypernymy Prediction via Taxonomy Enhanced Adversarial Learning. \textit{In Proceedings of the AAAI Conference on Artificial Intelligence (AAAI)}.

\smallskip \noindent
Samangouei, P.; Kabkab, M.; and Chellappa, R. 2018. Defense-GAN: Protecting classifiers against adversarial attacks using generative models. \textit{In International Conference on Learning Representations (ICLR)}.

\smallskip \noindent
Song, Y.; Kim, T.; Nowozin, S.; Ermon, S.; and Kushman, N. 2017. Pixeldefend: Leveraging generative models to understand and defend against adversarial examples. \textit{CoRR} abs/1710.10766, 2017.

\smallskip \noindent
Liao, F.; Liang, M.; Dong, Y.; and Pang, T. 2018. Defense against adversarial attacks using high-level representation guided denoiser. \textit{In IEEE Conference on Computer Vision and Pattern Recognition (CVPR)}.

\smallskip \noindent
Liu, C. R., Ye, D. P., and Shang, Y. Y. 2020. Defend sgainst sdversarial samples by using perceptual hash. \textit{Computers, Materials \& Continua} 62(3): 1365--1386.

\smallskip \noindent
Yang, P. Y. D.; Chen, J. B.; Hsieh, C. J.; Wang, J. L.; and Jordan, M. 2020.ML-LOO: Detecting Adversarial Examples with Feature Attribution. \textit{In Proceedings of the AAAI Conference on Artificial Intelligence (AAAI)}.

\smallskip \noindent
Cohen, J. M.; Rosenfeld, E.; and Kolter, J. Z. 2019. Certified adversarial robustness via randomized smoothing. \textit{CoRR} abs/1902.02918, 2019.

\smallskip \noindent
Li, B.; Chen, C. Y.; Wang, W. L.; and Carin, L. 2019. Certified adversarial robustness with additive noise. \textit{In the Conference on Neural Information Processing Systems (NeurIPS)}.

\smallskip \noindent
 Tao, G., Ma, S., Liu, Y., and Zhang, X. 2018. Attacks meet interpretability: Attribute steered detection of adversarial samples. \textit{In advances in Neural Information Processing Systems}, 7727--7738.

\smallskip \noindent
Simonyan, K.; Vedaldi, A.; and Zisserman, A. 2013. Deep inside convolutional networks: Visualising image classification models and saliency maps. \textit{CoRR} abs/1312.6034.

\smallskip \noindent
Springenberg, J. T.; Dosovitskiy, A.; Brox, T.; and Riedmiller, M. A. 2014. Striving for Simplicity: The All Convolutional Net. \textit{CoRR} abs/1412.6806.

\smallskip \noindent
Zeiler M. D.; and Fergus, R. 2014. Visualizing and understanding convolutional networks. \textit{In the European Conference on Computer Vision (ECCV)}.

\smallskip \noindent
Gan, C.; Wang, N.; Yang, Y.; Yeung, D. Y.; and Hauptmann, A. G. 2015. Devnet: A deep event network for multimedia event detection and evidence recounting. \textit{In IEEE Conference on Computer Vision and Pattern Recognition (CVPR)}.

\smallskip \noindent
Zhou, B.; Khosla, A.; Lapedriza, A.; Oliva, A.; and Torralba, A. 2016. Learning Deep Features for Discriminative Localization. \textit{In IEEE Conference on Computer Vision and Pattern Recognition (CVPR)}.

\smallskip \noindent
Selvaraju, R. R., Cogswell, M., Das, A., Vedantam, R., Parikh, D., and Betra, D. 2020. Grad-CAM: Visual Explanations from Deep Networks via Gradient-Based Localization. \textit{International Journal of Computer Vision} 128(2): 336--359.

\smallskip \noindent
Nicolae, M. I.; Sinn, M.; Tran, M. N.; Rawat, A.; Wistuba, M.; Zantedeschi, V.; Baracaldo, N.; Chen, B.; Ludwig, H.; Molly, I.M.; and Edwards, B. 2019. Adversarial Robustness Toolbox v0.4.0. \textit{CoRR} abs/1807.01069v3.

\end{document}